\documentclass[10pt,twocolumn,letterpaper]{article}

\usepackage{relsize}
\usepackage{iccv}
\usepackage{times}
\usepackage{epsfig}
\usepackage{graphicx}
\usepackage{amsmath}
\usepackage{amssymb}
\usepackage{mathdesign}
\usepackage{float}
\usepackage{rotating}
\usepackage{tabularx}
\usepackage{longtable}
\usepackage{multirow}
\usepackage{stringstrings}
\usepackage[export]{adjustbox}
\usepackage{setspace} 
\usepackage[linesnumbered,noend]{algorithm2e}
\usepackage[dvipsnames]{xcolor}
\usepackage{array}
\usepackage{booktabs}
\usepackage{enumitem}

\graphicspath{{./images/}{./}{./results/}}

\newcommand{\ignore}[1]{}

\renewcommand{\paragraph}[1]{\textbf{#1}}

\usepackage[pagebackref=true,breaklinks=true,letterpaper=true,colorlinks,bookmarks=false]{hyperref}

\iccvfinalcopy 

\def\iccvPaperID{266} 

\begin{document}

\title{Tracking as Online Decision-Making:\\ Learning a Policy from Streaming Videos with Reinforcement Learning}

\author{James Supan\v{c}i\v{c}, III\\
University of California, Irvine\\
{\tt\small jsupanci@uci.edu}
\and
Deva Ramanan\\
Carnegie Mellon University\\
{\tt\small deva@cs.cmu.edu}
}

\maketitle

\begin{abstract}
We formulate tracking as an online decision-making process, where a tracking agent must follow an object despite ambiguous image frames and a limited computational budget.
Crucially, the agent must decide where to look in the upcoming frames, when to reinitialize because it believes the target has been lost, and when to update its appearance model for the tracked object. 
Such decisions are typically made heuristically. 
Instead, we propose to learn an optimal decision-making policy by formulating tracking as a partially observable decision-making process (POMDP). 
We learn policies with deep reinforcement learning algorithms that need supervision (a reward signal) only when the track has gone awry. 
We demonstrate that sparse rewards allow us to quickly train on massive datasets, several orders of magnitude more than past work. 
Interestingly, by treating the data source of Internet videos as unlimited streams, we both learn and evaluate our trackers in a single, unified computational stream.
\end{abstract}

\section{Introduction}
\label{sec:intro}

\begin{figure}[tbp!]
\centering
\includegraphics[width=.8\linewidth]{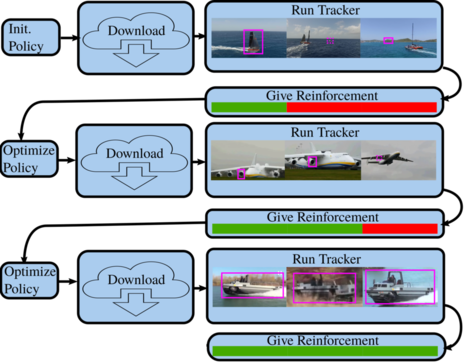}
\caption{\textbf{Streaming interactive training}: 
We propose an iterative procedure for interactively training trackers from data. 
We {\em download} a new video from the Internet and run the current
tracker on it, {\em evaluate} the tracker's performance with
interactive rewards, and then {\em retrain} the tracker policy (with reinforcement learning) with
the reward signals. Importantly, rather than requiring interactive labeling of bounding-boxes, we require only binary (incorrect / correct) feedback from human users. This scheme allows us to train and evaluate our tracker on massive streaming datasets, 100X larger than prior work (Table~\ref{tbl:datasets}).}
\label{fig:splash}
\label{fig:interactive_training_large}
\end{figure}

Object tracking is one of the basic computational building blocks of video analysis, relevant for tasks such as general scene understanding and perception-for-robotics. A particularly popular formalism is that of model-free tracking, where a tracker is provided with a bounding-box initialization of an unknown object. Much of the recent state-of-the-art advances make heavy use of machine learning~\cite{Kristan2016a,wang2015understanding,Pang2013,xiang2015learning,kahou2016ratm}, often producing impressive results by improving core components such as appearance descriptors or motion modeling. 

{\bf Challenges:} We see two significant challenges that limit further progress in model-free tracking. First, the {\em limited quantity of annotated video data} impedes both training and evaluation. While image datasets involve millions of images for training and testing, tracking datasets have hundreds of videos. Lack of data seems to arise from the difficulty of annotating videos, as opposed to images. Second, as vision (re)-integrates with robotics, video processing must be done in an online, streaming fashion. In terms of tracking, this requires a tracker to make {\em on-the-fly decisions} such as when to \emph{re-initialize} itself~\cite{Wang2015,hong2015multi,ma2015long,supancic2013self,kalal2012tracking} or
\emph{update} its appearance model (the so-called template-update problem~\cite{wang2015understanding,kalal2012tracking}). Such decisions are known to be crucial in terms of final performance, but are typically hand-designed rather than learned. 


{\bf Contribution 1 (interactive video processing):} We show that
reinforcement learning (RL) can be used to address both challenges in
distinct ways. In terms of data, rather than requiring videos to be
labeled with detailed bounding-boxes at each frame, we interactively
train trackers with far more limited supervision (specifying binary
rewards/penalties only when a tracker fails). This allows us to train
on massive video datasets that are {\bf $100 \times$ } larger than
prior work. Interestingly, RL also naturally lends itself to streaming
``open-world'' evaluation: when running a tracker on a
never-before-seen video, the video can be used for both evaluation of
the current tracker and for training (or refining) the tracker for
future use (Fig.~\ref{fig:splash}). This streaming evaluation allows
us to train and evaluate models in an integrated fashion
seamlessly. For completeness, we also evaluate our learned models on standard tracking benchmarks.

\begin{figure}[tbp!]
\centering
\includegraphics[width=.7\linewidth]{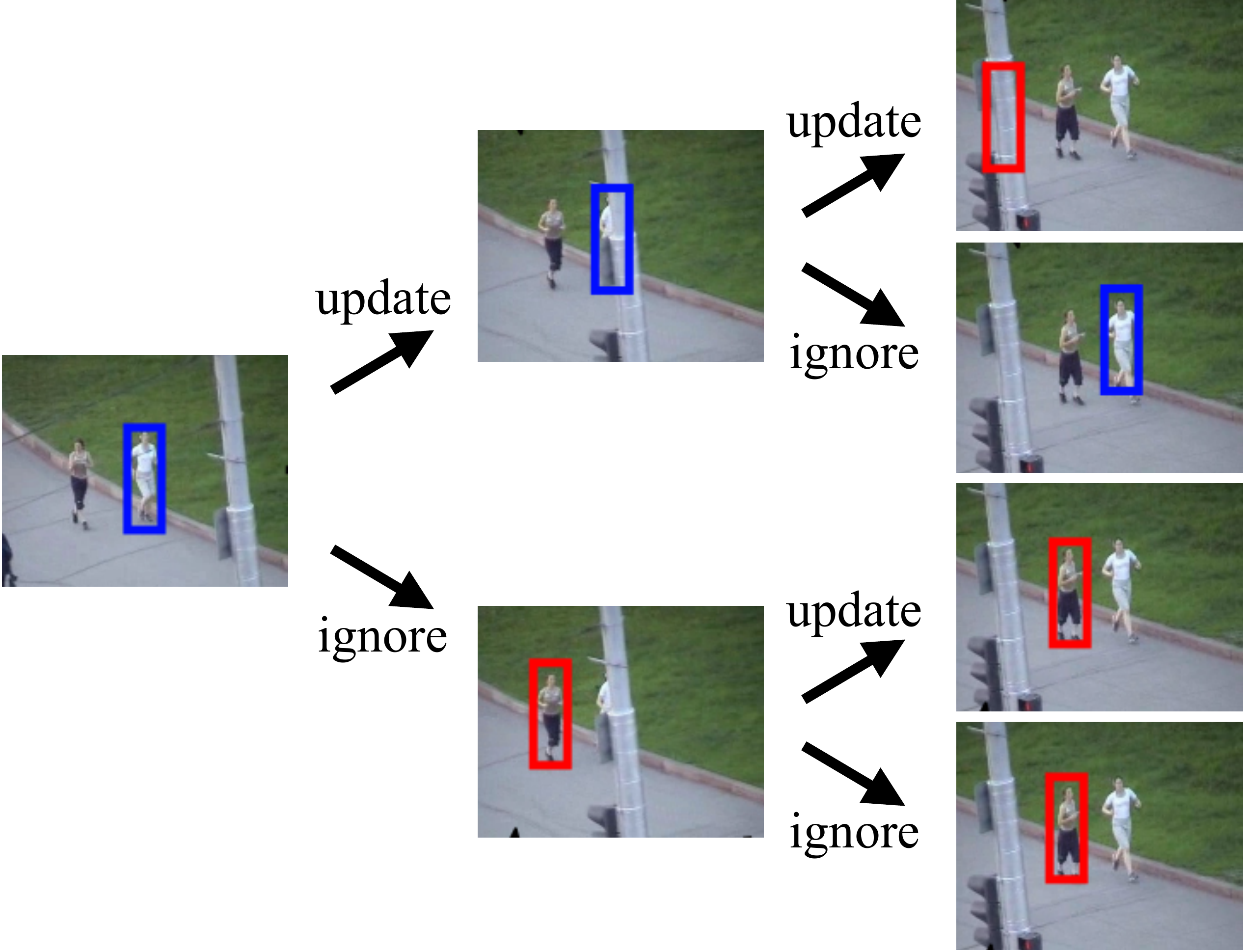}
\caption{\textbf{Decisions in tracking}: 
Trackers must decide when to update their appearance and when to re-initialize. This example enumerates the four possible outcomes of updating appearance (or not) over two  frames, where blue denotes a good track and red denotes an error.  Given a good track (left), it is important to update appearance to track through challenging frames with occlusions (center), but equally important to not update after an occlusion to prevent drift (right). Though such decisions are typically made heuristically, we recast tracking as a sequential decision-making process, and learn a policy with reinforcement learning. }
\label{fig:dec_tree}
\end{figure}

{\bf Contribution 2 (tracking as decision-making):} In terms of tracking, we model the tracker itself as an {\em active agent} that must make online decisions to maximize its reward, which is (as above) the correctness of a track. Decisions ultimately specify where to devote finite computational resources at any point of time: should the agent process only a limited region around the currently predicted location (e.g.,``track''), or should it globally search over the entire frame (``reinitialize'')? Should the agent use the predicted image region to update its appearance model for the object being tracked (``update''), or should it be ``ignored''? Such decisions are notoriously complicated when image evidence is ambiguous (due to say, partial occlusions): the agent may continue tracking an object but perhaps decide not to update its model of the object's appearance.
Rather than defining these decisions heuristically, we will ultimately use data-driven techniques to learn good {\em policies} for active decision-making (Fig.~\ref{fig:dec_tree}).

{\bf Contribution 3 (deep POMDPs):} 
We learn tracker decision policies using reinforcement learning.  
Much recent work in this space assumes a Markov Decision Process (MDP), where the agent observes the true state of the world~\cite{mnih2015human,xiang2015learning}, which is the true (possibly 3D) location and unoccluded appearance of the object being tracked.    
In contrast, our tracker only assumes that it receives {\em partial} image observations about the world state. The resulting partially-observable MDP (POMDP) violates Markov independence assumptions: actions depend on the entire history of observations rather than just the current one \cite{kaelbling1998planning,russell2003artificial}. 
As in \cite{hausknecht2015deep,khim2014qlearningtracking}, we account for this partial observability by maintaining a {\em memory} that captures {\em beliefs} about the world, which we update over time (Sec. \ref{sec:system}). In our case, beliefs capture object location and appearance, and action policies specify how and when to update those beliefs (e.g., how and when should the tracker update its appearance model)
(Sec. \ref{sec:interactive_training}). However, policy-learning is notoriously challenging because actions can have long-term effects on future beliefs. To efficiently learn policies, we introduce frame-based heuristics that provide strong clues as to the long-term effects of taking a particular action (Sec. \ref{sec:implementation}).

\section{Related Work}
\label{sec:related}

\paragraph{Tracking datasets:}
Several established benchmarks exist for evaluating trackers~\cite{wu2013online,Kristan2016a}.
Interestingly, there is evidence to suggest that many methods tend to overfit due to aggressive tuning~\cite{Pang2013}. 
Withholding test data annotation and providing an evaluation server addresses this to some extent~\cite{li2015nus,ILSVRC15}.
Alternatively, we propose to evaluate on an open-world stream of Internet videos, making over-fitting impossible by design. It is well-known that algorithms trained on ``closed-world'' datasets (say, with centered objects against clean backgrounds~\cite{palmer1981canonical,berg2009finding}) are difficult to generalize to ``in-the-wild'' footage~\cite{torralba2011unbiased}.
We invite the reader to compare our videos in the supplementary material to contemporary video benchmarks for tracking. 

\paragraph{Interactive tracking:} 
Several works have explored interactive methods that use trackers to help annotate data.
The computer first proposes a track.
Then a human corrects major errors and retrains the tracker using the
corrections~\cite{buchanan2006interactive,agarwala2004keyframe,vondrick2011video}.
Our approach is closely inspired by such active learning formalisms but differs in that we make use of minimal supervision in the form of a binary reward (rather than a bounding box annotation).

\paragraph{Learning-to-track: }
Many tracking benchmarks tend to focus on short-term tracking ($<2000$ frames per video)~\cite{wu2013online,Kristan2016a}. 
In this setting, a central issue appears to be modeling the appearance of the target. Methods that use deep features learned from large-scale training data perform particularly well~\cite{Wang2015,wangstct,ma2015hierarchical,Li2014,Nam2015}. 
\emph{Our focus on tracking over longer time frames poses additional challenges - namely, how to reinitialize after cuts, occlusions and failures, despite changes in target appearance~\cite{Maresca_2014_CVPR_Workshops,wang2015understanding}}. 
Several trackers address these challenges with hand-designed policies for model updating and reinitialization - TLD~\cite{kalal2012tracking}, ALIAN~\cite{pernici2014object} and SPL~\cite{supancic2013self} explicitly do so in the context of long-term tracking. On the other hand, our method takes a data-driven approach and learns policies for model-updating and reinitialization. Interestingly, such an ``end-to-end'' learning philosophy is often embraced by the multi-object tracking community, where strategies for online reinitialization and data association are learned from data~\cite{leal2015motchallenge,xiang2015learning,li2009learning}.
Most related to us are \cite{xiang2015learning}, who use an MDP for multi-object tracking, and \cite{khim2014qlearningtracking}, who use RL for single target tracking.
Both works use heuristics to reduce policy learning to a supervised learning task, avoiding the need to reason about rewards in the far future.
The robotics community has developed techniques to accelerate RL using human demonstrations~\cite{leon2013human}, interactive feedback~\cite{knox2011augmenting,thomaz2005real}, and hand-designed heuristics~\cite{bianchi2008accelerating}. 
Using heuristic functions for initialization~\cite{bianchi2008accelerating,knox2011augmenting}, our experimental results show that explicit Q-learning outperforms supervised reductions because it can learn to capture long-term effects of taking particular actions.

\paragraph{Real-time tracking through attention: } 
An interesting (but perhaps unsurprising) phenomenon is that better trackers tend to be slower~\cite{Kristan2016a}.
Indeed, on the VOT benchmark, most recent trackers do not run in real time.
Generally, trackers that search
locally~\cite{Montero_2015_ICCV_Workshops,vojir2013robust} run faster than those that search globally~\cite{supancic2013self,ma2015long,hua2015online}. 
To optimize visual recognition efficiency, one can learn a policy to guide selective search or attention.
Inspired by recent work which finds a policy for selective search using RL~\cite{kahou2016ratm,mnih2014recurrent,paletta2005q,chukoskie2013learning,jiang2016learning,bazzani2010learning,mathe2016reinforcement}, 
we also learn a policy that decides whether to track 
(\ie, search positions near the previous estimate) or reinitialize 
(\ie, search globally over the entire image).
But in contrast to this prior work, we additionally learn a policy to decide when to update a tracker's appearance model. 
To ensure that our tracker operates with a fixed computational budget, we implement reinitialization by searching over a random subset of positions (equal in number to those examined by  track). 

\begin{table}
\resizebox{\linewidth}{!}{
\begin{tabular}{lrrrl}
\hline \\
Dataset & \# Videos & \# Frames & Annotations & Type \\
\hline
OTB-2013~\cite{wu2013online} & 50            & 29,134      & 29,134  & AABB    \\
PTB~\cite{song2013tracking} & 50        & 21,551      & 21,551  & AABB    \\
VOT-2016~\cite{Kristan2016a} & 60   & 21,455      & 21,455  & RBB       \\
ALOV++~\cite{smeulders2014visual} & 315 & 151,657     & 30,331  & AABB      \\
NUS-PRO~\cite{li2015nus} & 365          & 135,310     & 135,310 & AABB     \\
\emph{Ours}              & 16,384       & 10,895,762  & 108,957 & Binary  \\
\hline
\end{tabular}}
\caption[Tracking datasets]{\textbf{Our interactive} learning formulation allows us to train and evaluate on dramatically more videos than prior work. 
We annotate binary rewards, while the other datasets provide Axis Aligned (AABB) or Rotated (RBB) Bounding Boxes. }
\label{tbl:datasets}
\end{table}

\begin{figure}[tbp!]
\centering
\includegraphics[width=1.0\linewidth]{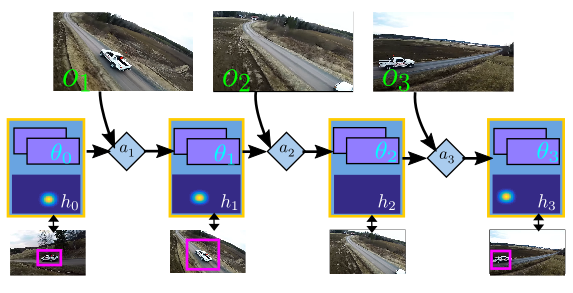}
\caption[Tracker architecture]{\textbf{Tracker architecture: }
At each frame $i$, our tracker updates a location heatmap $h_i$ for the target using the current image observation $o_i$, a location prior given by the previous frames' heatmap $h_{i-1}$, and the previous appearance model $\theta_{i-1}$. Crucially, our tracker learns a policy for actions $a_i$ that optimally update $h_i$ and $\theta_i$ \eqref{eq:update}.
}
\label{fig:tracker_arch}
\end{figure}

\section{POMDP Tracking}
\label{sec:system}

We now describe our POMDP tracker, using standard notation where possible~\cite{russell2003artificial}. For our purposes, a POMDP is defined by a tuple of states ($\Omega,O,A,B$): At each frame $i$, the world state $\omega_i \in \Omega$ generates a noisy observation $o_i \in O$ that is mapped by an agent into an action $a_i \in A$, which in turn generates a reward. 

In our case, the state $\omega_i$ captures the true location and appearance of the object being tracked in frame $i$. To help build intuition, one can think of the location as 2D pixel coordinates and appearance as a 2D visual template. Instead of directly observing this world state, the tracking agent maintains a belief over world states, written as $$b_i = (\theta_i,h_i), \quad \text{where} \quad \theta_i \in R^{h \times w \times f}, h_i \in R^{H \times W}$$ where $\theta_i$ is a distribution over appearances (we use a point-mass distribution encoded by a single $h \times w$ filter defined on $f$ convolutional features), and $h_i$ is a distribution over pixel positions (encoded as a spatial heatmap of size $H \times W$). Given the previous belief $b_{i-1}$ and current observed video frame $o_i$, the tracking agent updates its beliefs about the current frame $b_i$. Crucially, {\em tracker actions $a_i$ specify how to update its beliefs}, that is, whether to update the appearance model and whether to reinitialize by disregarding previous heatmaps. From this perspective, our POMDP tracker is a memory-based agent that learns a {\em policy} for when and how to update its own memory  (Fig.~\ref{fig:tracker_arch}). 

Specifically, beliefs are updated as follows:
\begin{align}
   h_i &= \left\{ \begin{array}{lr}
    TRACK(h_{i-1},\theta_{i-1},o_i) & \text{if} \quad  a_i^{(1)} = 1 \\
    REINIT(\theta_{i-1},o_i) & \quad \text{otherwise} \\
    \end{array}  \right. \\ \label{eq:update}
   \theta_i &= \left\{
   \begin{array}{lr}
   UPDATE(\theta_{i-1},h_i,o_i) & \text{if} \quad a_i^{(2)}  = 1\\
   \theta_{i-1} & \quad \text{otherwise} 
   \end{array}  \right. \nonumber
\end{align}
\noindent where  $a_i = (a_i^{(1)},a_i^{(2)})$.
Object heatmaps $h_i$ are updated by running the current appearance model $\theta_{i-1}$ on image regions $o_i$ near the target location previously predicted using $h_{i-1}$ (``tracking''). 
Alternatively, if the agent believes it has lost track,
it may globally evaluate its appearance model (``reinit'').
The agent may then decide to ``update'' its appearance model using the currently-predicted target location, or it may leave its appearance model unchanged.
In our framework, the tracking, reinitialization, and appearance-update modules can be treated as black boxes given the above functional form. 
We will discuss particular implementations shortly, but first, we focus on the heart of our RL approach: a principled framework for learning to take appropriate actions. To do so, we begin by reviewing standard approaches for decision-making.

{\bf Online heuristics:} The simplest approach to picking actions might be to pre-define heuristics functions which estimate the correct action. For example, many trackers reinitialize whenever the maximum confidence of the heatmap is below a threshold. Let us summarize the information available to the tracker at frame $i$ as a ``state'' $s_i$, which includes the previous belief $b_{i-1}$ (the previous heatmap and appearance model) and current image observation $o_i$. 
\begin{align}
    a_i^* = \text{Heur}_{on}(s_i), \quad \text{where} \quad s_i = (b_{i-1},o_i)
\end{align}

{\bf Offline heuristics:} A generalization of the above is to use {\em offline} training data to build better heuristics. Crucially, one can now make use of ground-truth training annotations as well as future frames to better gauge the impact of possible actions. We can write this knowledge as the true world state $\{\omega_i, \forall i\}$. 
For example, a natural heuristic may be to reinitialize whenever the predicted object location does not overlap the ground-truth position for that frame. Similarly, one may update appearance whenever doing so improves the confidence of ground-truth object locations across future frames:
\begin{align}
    a_i^* = \text{Heur}_{off}(s_i,\{\omega_i: \forall i\}) \label{eq:heur}
\end{align}
Crucially, these heuristics cannot be applied at test time because ground truth is not known! However, they can generate per-frame target action labels $a_i^{*}$ on training data, effectively reducing policy learning to a standard {\em supervised-learning problem}. Though offline heuristics appear to be a simple and intuitive approach to policy learning, we have not seen them widely applied for learning tracker action policies. 

{\bf Q-functions:} Offline heuristics can be improved by unrolling them forward in time: the benefit of a putative action can be better modeled by applying that action, processing the next frame, and using the heuristic to score the ``goodness'' of possible actions in that next frame. This intuition is formalized through the well-known {\em Bellman equations} that recursively define Q-functions to return a goodness score (the expected future reward) for each putative action $a$:
\begin{align}
        Q(s_i,a_i) = R(s_i) + \gamma \max_{a_{i+1}}Q(s_{i+1},a_{i+1}), 
\end{align}
\noindent where $s_i$ includes both the tracker belief state and image observation, and $R(s_i)$ is the reward associated with the reporting the estimated object heatmap $h_i$. We let $R(s_i)=1$ for a correct prediction and $0$ otherwise. 
Finally, $\gamma \in [0,1]$ is a discount factor that trades off immediate vs. future per-frame rewards. 
Given a tracker state and image observation $s_i$, the optimal action to take is readily computed from the Q-function: $$a^* = \arg\max_a Q(s_i,a)$$

{\bf Q-learning:} Traditionally, Q-functions are iteratively learned with Q-learning~\cite{russell2003artificial}:
\begin{align}
    Q(s_i,a_i) \Leftarrow &Q(s_i,a_i) + \ldots\\
    &\alpha \Big( R(s_i) + \gamma \max_{a_{i+1}} Q(s_{i+1},a_{i+1}) - Q(s_i,a_i) \Big) \nonumber
\end{align}
\noindent where $\alpha$ is a learning rate. 
To handle continuous belief states, we approximate the Q-function with a CNNs: $$Q(s_i,a_i) \approx CNN(s_i,a_i)$$ that processes states $s_i$ and binary actions $a_i$ to return a scalar value capturing the expected future reward for taking that action.  
Recall that a state $s_i$ encodes a heatmap and an appearance model from previous frames and an image observation from the current frame. 

\section{Interactive Training and Evaluation}
\label{sec:interactive_training}

\newcommand\mycommfont[1]{\footnotesize\ttfamily\textcolor{OliveGreen}{#1}}
\SetCommentSty{mycommfont}
\begin{algorithm}[tbp!]
\While{True}{
\text{Download random video}\;
    $\theta_1 \leftarrow \text{UPDATE}  (h_1, o_1)$; \tcc{manually init.}\
    \ForAll{$i \in \text{video}$}{ 
        $s_i = ((\theta_{i-1}, h_{i-1}), o_i)$\;
        \tcc{\textbf{t}rack or \textbf{r}einitialize?}
        \If{$Q_w(s_i, \textbf{t}) > Q_w(s_i, \textbf{r})$}{
            $h_i \leftarrow \text{TRACK}(h_{i-1}, \theta_{i-1}, o_{i})$; $a_i^{(1)} \leftarrow 1$\;
        }\lElse{
            $h_i \leftarrow \text{REINIT}(\theta_{i-1},o_i)$; $a_i^{(1)} \leftarrow 0$
        }
        \tcc{\textbf{u}pdate or \textbf{i}gnore? }
        \If{$Q_w(s_i, \textbf{m}) > Q_w(s_i, \textbf{i})$}{
            $\theta_i \leftarrow \text{UPDATE}(\theta_{i-1},h_i,o_i)$; $a_i^{(2)} \leftarrow 1$\;
        }\lElse{
            $\theta_i \leftarrow \theta_{i-1}$ ;$a_i^{(2)} \leftarrow 0$
        }
    \tcc{manually evaluate the performance}
     $r_i \leftarrow \text{annotated frame correct?}$ \;
    \tcc{update experience database}
    $\mathcal{D} \mathsmaller{\leftarrow} \mathcal{D} \cup (s_i,a_i,r_i,s_{i+1})$
    }
    $w \leftarrow \arg\min L(w;\mathcal{D})$
}
\caption[Reinforcement learning for long-term tracking]{Our final learning algorithm interactively labels a streaming dataset of videos while learning a tracker action policy $Q_w$. Given a video, steps 5 through 11 run a tracker according to the current policy (visualized in detail in Fig.~\ref{fig:runtimeflowbig}). An annotator then assesses the binary reward $r_i$ (correctness) for the highest-scoring bounding box extracted from the heatmap $h_i$ by using an intersection-over-union threshold. Annotated frames (and their associated state-action-reward-nextstate tuples) are added to our experience replay database $D$. We then sample a minibatch of replay memories and update the action policy $w$ with backprop (Eq.~\ref{eq:q}).}
\label{alg:drl}
\end{algorithm}

\begin{figure*}
    \centering
    \includegraphics[width=.8\linewidth,trim = 10cm 0mm 0mm 0mm, clip=true]{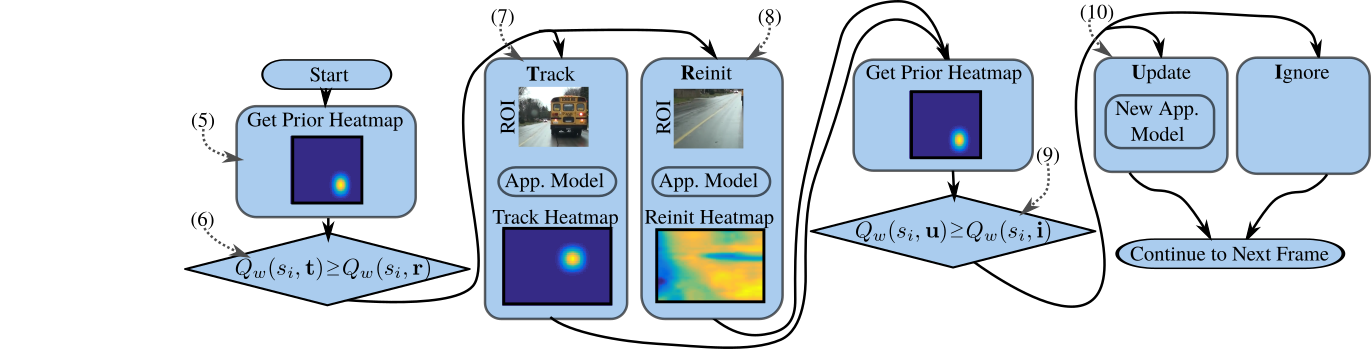}
   \caption[Our p-tracker's test time pipeline]{\textbf{Our tracker's
       (p-track's)} test time pipeline is shown above as a standard
     flow chart, with correspondences to Alg.~\ref{alg:drl} numbered.
     The tracker invokes this pipeline once per frame. }
    \label{fig:runtimeflowbig}
\end{figure*}

In this section, we describe our procedure for interactively learning CNN parameters $w$ (that encode tracker action policies) from streaming video datasets. To do so, we gradually build up a database of {\em experience replay memories}~\cite{adam2012experience,mnih2015human}, which are a collection of state-action-reward-nextstate tuples $\mathcal{D} = \{(s_i,a_i,r_i,s_{i+1})\}$.
Q-learning reduces policy learning to a {\em supervised regression
  problem} by unrolling the current policy one step forward in time.
This unrolling results in the standard Q-loss
function~\cite{mnih2015human}: 
\begin{align}
    L(w,\mathcal{D}) = \Big( r_i + \max_{a_{i+1}} Q_w(s_{i+1},a_{i+1}) - Q_w(s_i,a_i) \Big)^2 \label{eq:q}
\end{align}
Gradient descent on the above objective is performed as follows: given a training sample $(s_i,a_i,r_i,s_{i+1})$, 
first perform a forward pass to compute the current estimate $Q_w(s_i,a_i)$ and 
the target: $r_i + \max_{a_{i+1}} Q_w(s_{i+1},a_{i+1})$. 
Then backpropagate through the weights $w$ to reduce $L(w)$. 


The complete training algorithm is written in Alg.~\ref{alg:drl} and illustrated in Fig.~\ref{fig:interactive_training_large}.
We choose random videos from the Internet 
by sampling phrases using WordNet~\cite{miller1995wordnet}.  Given the sampled phrase and video, an annotator provides an
initialization bounding box and begins running the existing tracker. After tracking, the annotator marks those frames (in strides of 50) where the tracker was incorrect using a standard 50\% intersection-over-union threshold. 
Such binary annotation (``correct'' or ``failed'') requires far less time per frame than bounding-box annotation: we design a real-time interface that simply requires a user to depress a button during tracker failures. By playing back videos at a (user-selected) sped-up frame rate, users annotate 1200 frames per minute on average (versus 34 for bounding boxes). Annotating our entire dataset of 10 million frames (Table~\ref{tbl:datasets}) requires a little under two-days of labor, versus the months required for equivalent bounding-box annotation. After running our tracker and interactively marking failures, we use the annotation as a reward signal to update the policy parameters for
the next video. Thus each video is used to both evaluate the current
tracker and train it for future videos. 

\section{Implementation}
\label{sec:implementation}
In this section, we discuss several implementation details. We begin with a detailed overview of our tracker's test time pipeline. The implementation details for the TRACK, REINIT, UPDATE, and Q-functions follow. Finally, we describe the offline heuristics used at train time. 
\label{subsec:tracker_design}

\paragraph{Overview of p-tracker's pipeline: } In
Fig.~\ref{fig:runtimeflowbig} we show a flowchart for our tracker's
test time pipeline. We use a UML Activity
Diagram~\cite{rumbaugh2004unified}. 
We now describe how we chose to implement the TRACK, REINIT, UPDATE and Q-functions.

\paragraph{TRACK/REINIT functions:} Our TRACK function
(Eq.~\ref{eq:update}) takes as input the previous heatmap $h_{i-1}$,
appearance model $\theta_{i-1}$, and image observation $o_i$, and
produces a new heatmap for the current frame. We experiment with
implementing TRACK using either of two state-of-the-art fully-convolutional trackers: FCNT
\cite{Wang2015} or CCOT~\cite{danelljan2016beyond}.
We refer to the reader to the original works for precise implementation details, but summarize them here: TRACK crops the current image $o_i$ to a region of interest (ROI) around the most likely object location from the previous frame (the argmax of $h_{i-1}$). This ROI is resized to a canonical image size (e.g., $224 \times 224$) and processed with a CNN (VGG~\cite{Simonyan14c}) to produce a convolutional feature map. The object appearance model $\theta_{i-1}$ is represented as a filter on this feature map, allowing one to compute a new heatmap with a convolution. 
When the tracker decides that it has lost track, the REINIT model simply processes a random ROI. In general, we find CCOT to outperform FCNT (consistent with the reported literature) and so focus on the former in our experiments, though we include diagnostic results with both trackers (to illustrate the generality of our approach).

\begin{figure}[tbp!]
\centering
\includegraphics[width=.7\linewidth]{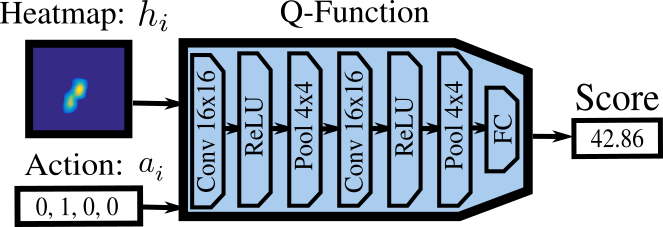}
\caption[Implementing $Q$ as a CNN]{\textbf{$Q$-CNNs: } 
A Q-function predicts a score (the expected future reward) as a
function of (1) the localization heatmap and (2) an action encoded
using a one-hot encoding. We implement our $Q$-functions using the architecture shown
above: two convolutional layers followed by a fully-connected (FC) layer. }
\label{fig:appendix_decision}
\label{fig:decision}
\end{figure}

{\bf UPDATE function:} We update the current filter $\theta$ using
positive and negative patches extracted from the current frame $i$. We
extract a positive patch from the maximal location in the reported
heatmap $h_i$, and extract negative patches from adjacent regions with
less than 30\% overlap. We update $\theta$ following the default scheme in the underlying tracker:
for FCNT, $\theta$ is a two-layer convolutional template that is updated with a fixed number of gradient descent iterations (10). For CCOT, $\theta$ is a multi-resolution set of
convolutional templates that is fit through conjugate gradients.  

{\bf Q-function CNN:} Recall that our Q-functions process a tracker state $s_i = ((h_{i-1},\theta_{i-1}),o_i)$ and a candidate action $a_i$, to return a scalar representing the expected future reward of taking that action (Fig.~\ref{fig:decision} and Eq.~\ref{eq:q}).
In practice, we define two separate Q-functions for our two binary decisions (TRACK/REINIT and UPDATE/IGNORE). To plug into standard learning approaches, we formally define a single $Q$ function as the sum of the two functions, implying that the optimal decisions can be computed independently for each. 
We found it sufficed to condition on the heatmap $h_{i-1}$ and
implemented each function as a CNN, where the first two hidden layers are shared across the two functions. 
Each shared hidden layer consists of $16 \times 16 \times 4$ convolution followed by ReLU and $4 \times 4$ max pooling. For each decision, an independent fully-connected layer ultimately predicts the expected future reward. 
When training the Q-function using experience-replay, we use $\gamma = .95$, a learning rate of 1e-4, a momentum of .9 and 1e-8 weight decay.

{\bf Offline heuristics:}
Deep q-learning is known to be unstable, and we found good initialization was important for reliable convergence. We initialize the Q-functions in Eq.~\ref{eq:q} (which specify the goodness of particular actions) with the offline heuristics from Eq.~\ref{eq:heur}. Specifically, the heuristic action $a_i^*$ has a goodness of 1 (scaled by future discount rewards), while other actions have a goodness of 0:
\begin{align}
    Q_{\text{init}}(s_i,a_i) \Leftarrow {\bf I}[a_i = a_i^*] \sum_{j \geq i} \gamma^{j-i} \label{eq:qinit}
\end{align}
\noindent where ${\bf I}$ denotes the identity function. In practice, we found it useful to minimize a weighted average of the true loss in Eq.~\ref{eq:q} and a supervised loss  $(Q_w - Q_{\text{init}})^2$, an approach related to heuristically-guided Q-learning ~\cite{kahou2016ratm,bianchi2012heuristically}.
Defining a heuristic for TRACK vs REINIT is straightforward: $a^*_i$ should TRACK whenever the peak of the reported heatmap overlaps the ground-truth object on frame $i$. 
Defining a heuristic for UPDATE vs IGNORE is more subtle. Intuitively, $a^*_i$ should UPDATE appearance with frame $i$ whenever doing so improves the confidence of future ground-truth object locations in that video. 
To operationalize this, we update the current appearance model $\theta_i$ on samples from frame $i$ and compute $\Delta_+$,  the number of future frames where the updated appearance increases confidence of ground-truth locations (and similarly $\Delta_-$, the number of frames where the update decreases confidence of track errors). 
We set $a^*_i$ to update when $\Delta_+ + \Delta_- >.5N$, where N is the total number of future frames.

\section{Experiments}
\label{sec:experiments}

\begin{figure*}[tbp!]
\centering
\bgroup
\renewcommand{\tabcolsep}{0pt}
\def\arraystretch{0}
\begin{tabular}{c|c|c|c|c|c|}
\multicolumn{6}{c}{Online Tracking Benchmark~\cite{wu2013online}} \\
\includegraphics[width=.16\linewidth,height=.125\linewidth]{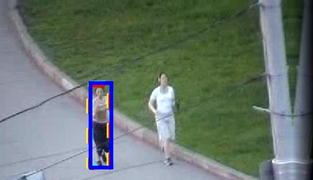} & 
\includegraphics[width=.16\linewidth,height=.125\linewidth]{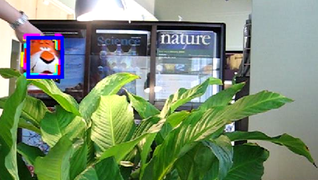} & 
\includegraphics[width=.16\linewidth,height=.125\linewidth]{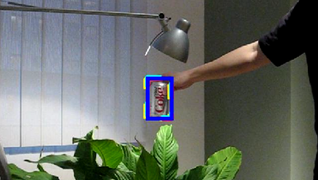} &
\includegraphics[width=.16\linewidth,height=.125\linewidth]{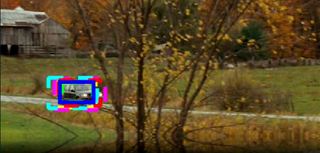} &
\includegraphics[width=.16\linewidth,height=.125\linewidth]{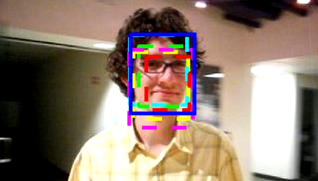} &
\includegraphics[width=.16\linewidth,height=.125\linewidth]{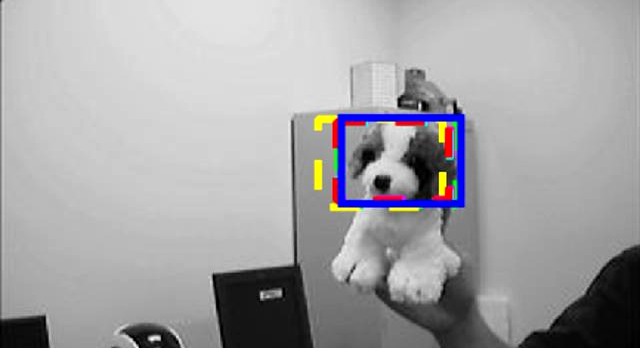} \\
\includegraphics[width=.16\linewidth,height=.125\linewidth]{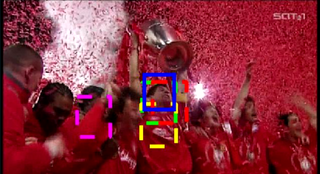} & 
\includegraphics[width=.16\linewidth,height=.125\linewidth]{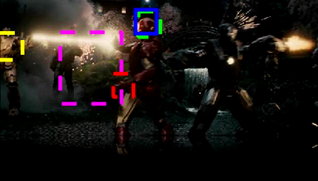} &
\includegraphics[width=.16\linewidth,height=.125\linewidth]{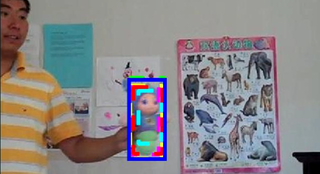}&
\includegraphics[width=.16\linewidth,height=.125\linewidth]{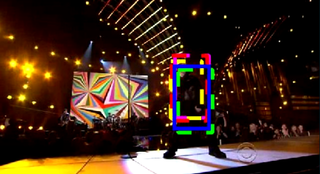} &
\includegraphics[width=.16\linewidth,height=.125\linewidth]{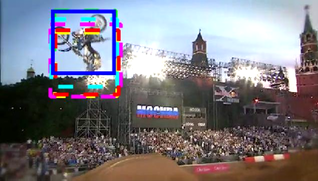} &
\includegraphics[width=.16\linewidth,height=.125\linewidth]{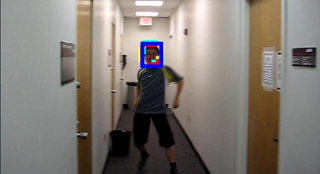}
\end{tabular}
\egroup
\caption[OTB-2013 dataset]{\textbf{OTB-2013: } Our \textcolor{blue}{p-tracker} (solid line) compared to
\textcolor{green}{FCNT}~\cite{Wang2015}, 
\textcolor{red}{MUSTer}~\cite{hong2015multi}, 
\textcolor{yellow}{LTCT}~\cite{ma2015long}, 
\textcolor{cyan}{TLD}~\cite{kalal2012tracking} and 
\textcolor{magenta}{SPL}~\cite{supancic2013self} (dashed lines). In general, many videos are easy for modern trackers, implying that a method's rank is determined by a few challenging videos (such as the confetti celebration and fireworks on the bottom left). Our tracker learns an UPDATE and REINIT policy that does well on such videos.}
\label{fig:qual_otb}
\end{figure*}

\begin{figure}[tbp!]
  \centering
\includegraphics[width=\linewidth]{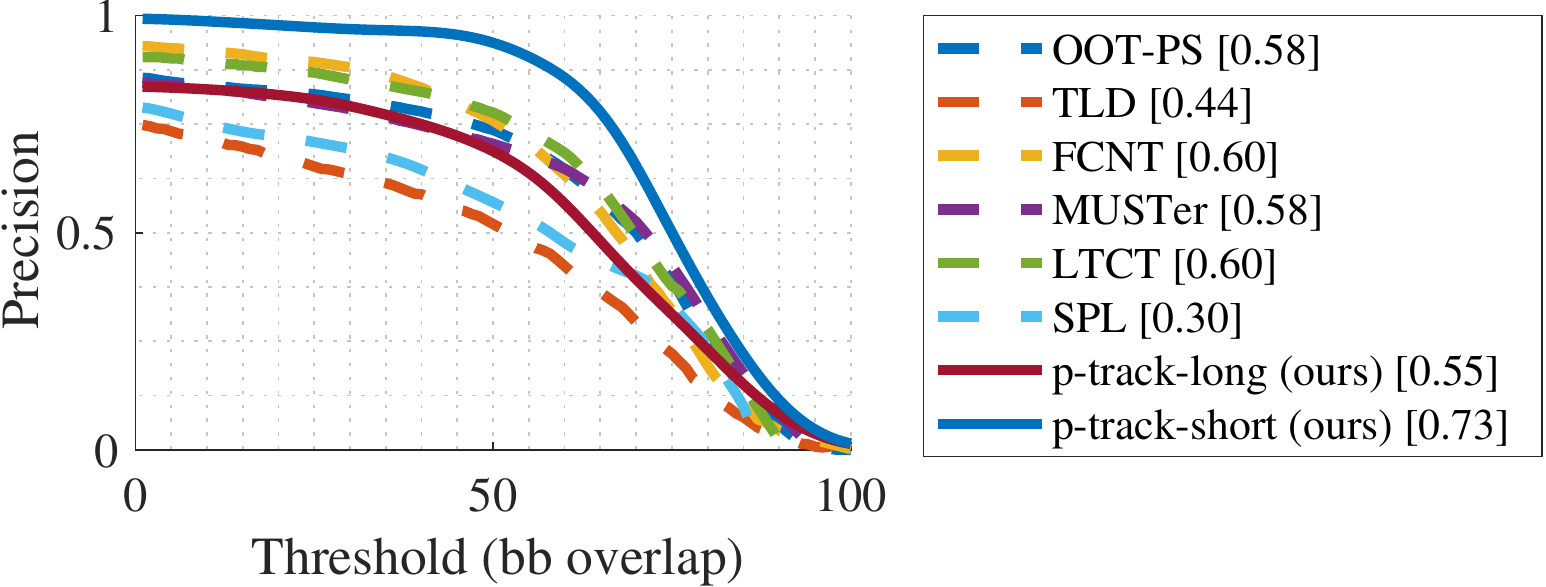} 
\caption[OTB-2013 results]{\textbf{OTB-2013~\cite{wu2013online} results}: 
Our learned policy tracker (p-track) performs competitively on standard short-term tracking benchmarks. We find that a policy learned for long-term tracking (p-track-long) tends to select the IGNORE action more often (appropriate during occlusions, which tend be more common in long videos). Learning a policy from short-term videos significantly improves performance, producing state-of-the-art results: compare our p-track-short vs. OOT-PS~\cite{hua2015online}, TLD~\cite{kalal2012tracking}, FCNT~\cite{Wang2015}, MUSTer~\cite{hong2015multi}, LTCT~\cite{ma2015long}, and SPL~\cite{supancic2013self}. }
\label{fig:benchmarks}
\end{figure}

\begin{figure}[tbp!]
        \centering
        \includegraphics[width=.90\linewidth]{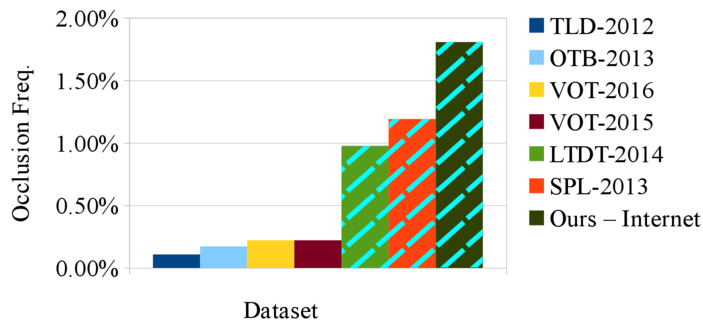}
        \caption[Occlusion Frequency]{\textbf{Occlusion Frequency: } We compare how frequently targets become occluded ($\frac{\text{\# occlusions}}{\text{\# frames}}$) on various short-term (left, solid) and long-term tracking datasets (right, hatched). Long-term datasets contain more frequent occlusions. To avoid drifting during these occlusions, trackers need to judiciously decide when to update their templates and may need to reinitialize more often.  This motivates the need for formally {\em learning} decision policies tailored for different tracking scenarios, rather than using a single set of hard-coded heuristics.}
        \label{fig:reinit_freq}
\end{figure}

\begin{figure}[tbp!]
  \centering
    \raisebox{-.5\height}{\includegraphics[width=.38\linewidth]{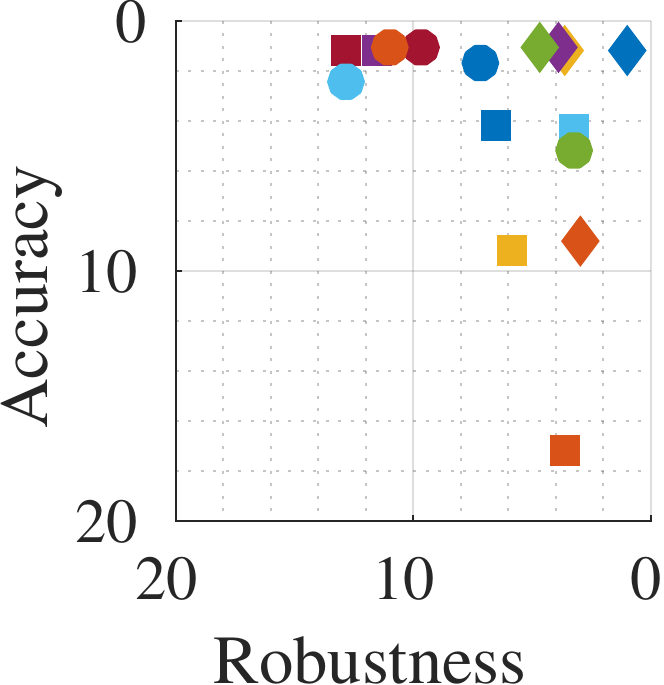}} 
    \resizebox{.26\textwidth}{!}{
        \begin{tabular}{ll}
{\color[rgb]{0.000000,0.447000,0.741000} $\blacklozenge$} p-track(ours)  & {\color[rgb]{0.850000,0.325000,0.098000} $\blacklozenge$} FCNT \\ 
{\color[rgb]{0.929000,0.694000,0.125000} $\blacklozenge$} CCOT~\cite{danelljan2016beyond}  & {\color[rgb]{0.494000,0.184000,0.556000} $\blacklozenge$} TCNN~\cite{nam2016modeling} \\ 
{\color[rgb]{0.466000,0.674000,0.188000} $\blacklozenge$} SSAT~\cite{nam2016learning}  & {\color[rgb]{0.301000,0.745000,0.933000} $\blacksquare$} MLDF~\cite{wang2016stct} \\ 
{\color[rgb]{0.635000,0.078000,0.184000} $\blacksquare$} Staple~\cite{bertinetto2016staple}  & {\color[rgb]{0.000000,0.447000,0.741000} $\blacksquare$} DDC~\cite{Kristan2016a} \\ 
{\color[rgb]{0.850000,0.325000,0.098000} $\blacksquare$} EBT~\cite{zhu2016beyond}  & {\color[rgb]{0.929000,0.694000,0.125000} $\blacksquare$} SRBT~\cite{Kristan2016a} \\ 
{\color[rgb]{0.494000,0.184000,0.556000} $\blacksquare$} STAPLEp~\cite{bertinetto2016staple}  & {\color[rgb]{0.466000,0.674000,0.188000} $\bullet$} DNT~\cite{chi2017dual} \\ 
{\color[rgb]{0.301000,0.745000,0.933000} $\bullet$} SSKCF~\cite{lee2011visual}  & {\color[rgb]{0.635000,0.078000,0.184000} $\bullet$} SiamRN~\cite{bertinetto2016fully} \\ 
{\color[rgb]{0.000000,0.447000,0.741000} $\bullet$} DeepSRDCF~\cite{danelljan2015learning}  & {\color[rgb]{0.850000,0.325000,0.098000} $\bullet$} SHCT~\cite{du2016online} \\ 
    \end{tabular}}
    \vspace{2pt}
    \caption[VOT-2016 results]{\textbf{VOT-2016~\cite{Kristan2016a}
        results}: Our learned policy tracker (p-track-short) is as
      accurate as the state-of-the-art but is considerably more
      robust. Robustness is measured by a ranking of trackers
      according to the number of times they fail, while accuracy is
      the rank of a tracker according to its average overlap with the
      ground truth. Notably, p-track significantly outperforms FCNT~\cite{Wang2015} and  CCOT~\cite{danelljan2016beyond} in terms of robustness, even though its TRACK and UPDATE modules follow directly from those works. }
    \label{fig:voteval}
\end{figure}

\paragraph{Evaluation metrics:} Following established protocols for long-term tracking~\cite{kalal2012tracking,supancic2013self}, 
we evaluate $F1 =\frac{2pr}{p+r}$, where precision $(p)$ is the fraction of predicted locations that are correct and recall $(r)$ is the fraction of ground-truth locations that are correctly predicted.
Because Internet videos vary widely in difficulty, we supplement averages with boxplots to better visualize performance skew. When evaluating results on standard benchmarks, we use the default evaluation criteria for that benchmark.



\begin{figure*}[tbp!]
\centering
\resizebox{.9\linewidth}{!}{
\bgroup
\renewcommand{\tabcolsep}{0pt}
\def\arraystretch{0}
\begin{tabular}{c|c|c|c}
\multicolumn{4}{c}{Internet Videos}  \\
\rotatebox[origin=c]{90}{\small{Transform}}&
\raisebox{-.5\height}{\includegraphics[width=.24\linewidth,height=.13\linewidth]{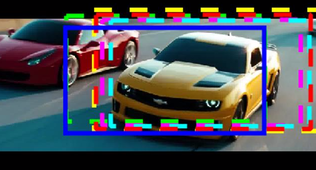}} & 
\raisebox{-.5\height}{\includegraphics[width=.24\linewidth,height=.13\linewidth]{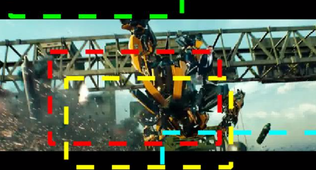}} & 
\raisebox{-.5\height}{\includegraphics[width=.24\linewidth,height=.13\linewidth]{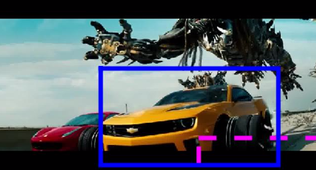}} \\
\rotatebox[origin=c]{90}{\small{SpaceX}}&
\raisebox{-.5\height}{\includegraphics[width=.24\linewidth,height=.13\linewidth]{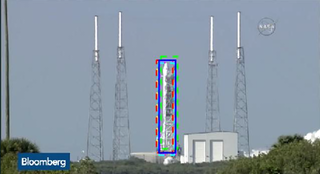}} & 
\raisebox{-.5\height}{\includegraphics[width=.24\linewidth,height=.13\linewidth]{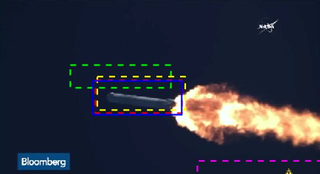}} & 
\raisebox{-.5\height}{\includegraphics[width=.24\linewidth,height=.13\linewidth]{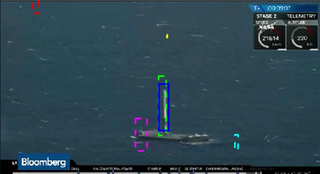}} \\
\rotatebox[origin=c]{90}{\small{SnowTank}}&
\raisebox{-.5\height}{\includegraphics[width=.24\linewidth,height=.13\linewidth]{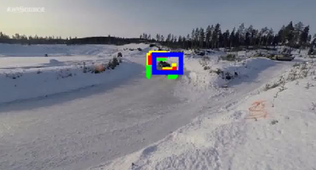}} & 
\raisebox{-.5\height}{\includegraphics[width=.24\linewidth,height=.13\linewidth]{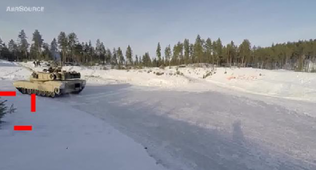}} & 
\raisebox{-.5\height}{\includegraphics[width=.24\linewidth,height=.13\linewidth]{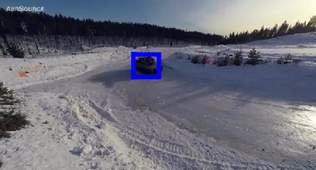}} \\
\rotatebox[origin=c]{90}{\small{JohnWick}}&
\raisebox{-.5\height}{\includegraphics[width=.24\linewidth,height=.13\linewidth]{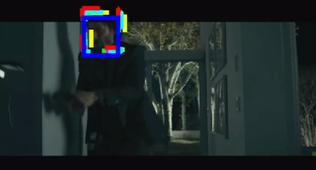}} & 
\raisebox{-.5\height}{\includegraphics[width=.24\linewidth,height=.13\linewidth]{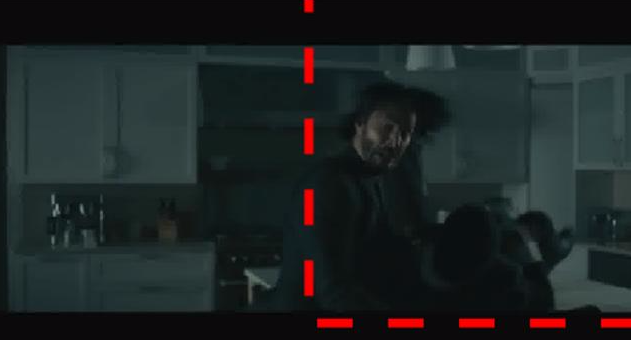}} & 
\raisebox{-.5\height}{\includegraphics[width=.24\linewidth,height=.13\linewidth]{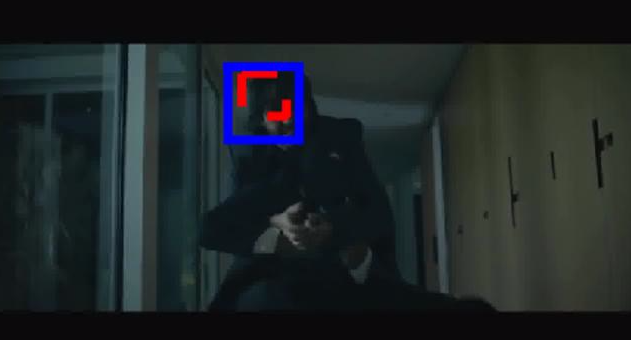}} \\
\end{tabular}
\egroup}
\caption[New Internet dataset]{\textbf{Internet videos} contain new challenges, such as cuts, 
strange and interesting behaviors, fast motion and complex illumination. 
Our \textcolor{blue}{p-tracker} (solid line) learns a policy that outperforms \textcolor{green}{FCNT}~\cite{Wang2015}, 
\textcolor{red}{MUSTer}~\cite{hong2015multi}, 
\textcolor{yellow}{LTCT}~\cite{ma2015long}, 
\textcolor{cyan}{TLD}~\cite{kalal2012tracking} and 
\textcolor{magenta}{SPL}~\cite{supancic2013self} (dashed lines).}
\label{fig:qual}
\end{figure*}

\begin{figure}[tbp!]
\hspace{80pt} {\small F1-measure}
   \centering
    \includegraphics[width=.95\linewidth,trim = 0mm 10mm 0mm 0mm, clip=true]{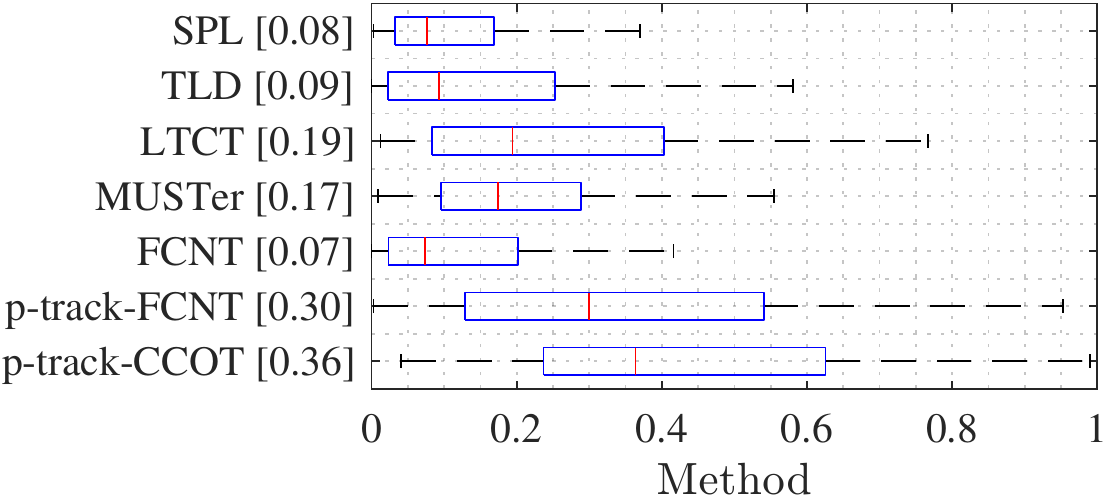}
    \caption[Results on Internet videos]{\textbf{Versus state-of-the-art}: Our learned policy (\textit{p-track}) performs better than state-of-the-art baselines~\cite{Wang2015,hong2015multi,ma2015long,kalal2012tracking,supancic2013self} on a held-out test set. See Sec.~\ref{sec:learn2track_comparative_eval} for discussion.  
     }
    \label{fig:comparative}
\end{figure}

\subsection{Comparative Evaluation}
\label{sec:learn2track_comparative_eval}

\paragraph{Short-term benchmarks: } 
While our focus is long-term tracking, we begin by presenting results on existing benchmarks that tend to focus on the short-term setting -- the Online Tracker Benchmark (OTB-2013)~\cite{wu2013online} and Visual Object Tracking Benchmark (VOT-2016)~\cite{Kristan2016a}. 
Our \textbf{p}olicy \textbf{track}er (\emph{p-track-long}), trained on Internet videos, performs competitively (Fig.~\ref{fig:benchmarks}), but tends to over-predict occlusions
(which rarely occur in short-term videos, as shown in Fig.~\ref{fig:reinit_freq}).
Fortunately, we can learn dedicated policies for short-term tracking (\emph{p-track-short}) by applying reinforcement learning (Alg.~\ref{alg:drl}) on short-term training videos.
For each test video in OTB-2013, we learn a policy using the 40 most dissimilar videos in VOT-2016 (and vice-versa). We define similarity between videos to be the correlation between the average (ground-truth) object image in RGB space.
This ensures that, for example, we do not train using \emph{Tiger1} when testing on \emph{Tiger2}. 
Even under this controlled scenario, \emph{p-track-short} significantly outperforms prior work on both OTB-2013 (Figs.~\ref{fig:qual_otb} and~\ref{fig:benchmarks}) and VOT-2016 (Fig.~\ref{fig:voteval}). 

\paragraph{Long-term baselines:} For the long-term setting, we compare to two classic long-term trackers: TLD~\cite{kalal2012tracking} and SPL~\cite{supancic2013self}. Additionally, we also compare against short-term trackers with public code that we were able to adapt: FCNT~\cite{Wang2015}, MUSTer~\cite{hong2015multi}, and LTCT~\cite{ma2015long}. Notably, all these baselines use hand-designed heuristics for deciding when to appearance update and reinitialize.

\begin{figure}[tbp!]
\centering
\includegraphics[width=.9\linewidth,trim={0 0 0 0},clip]{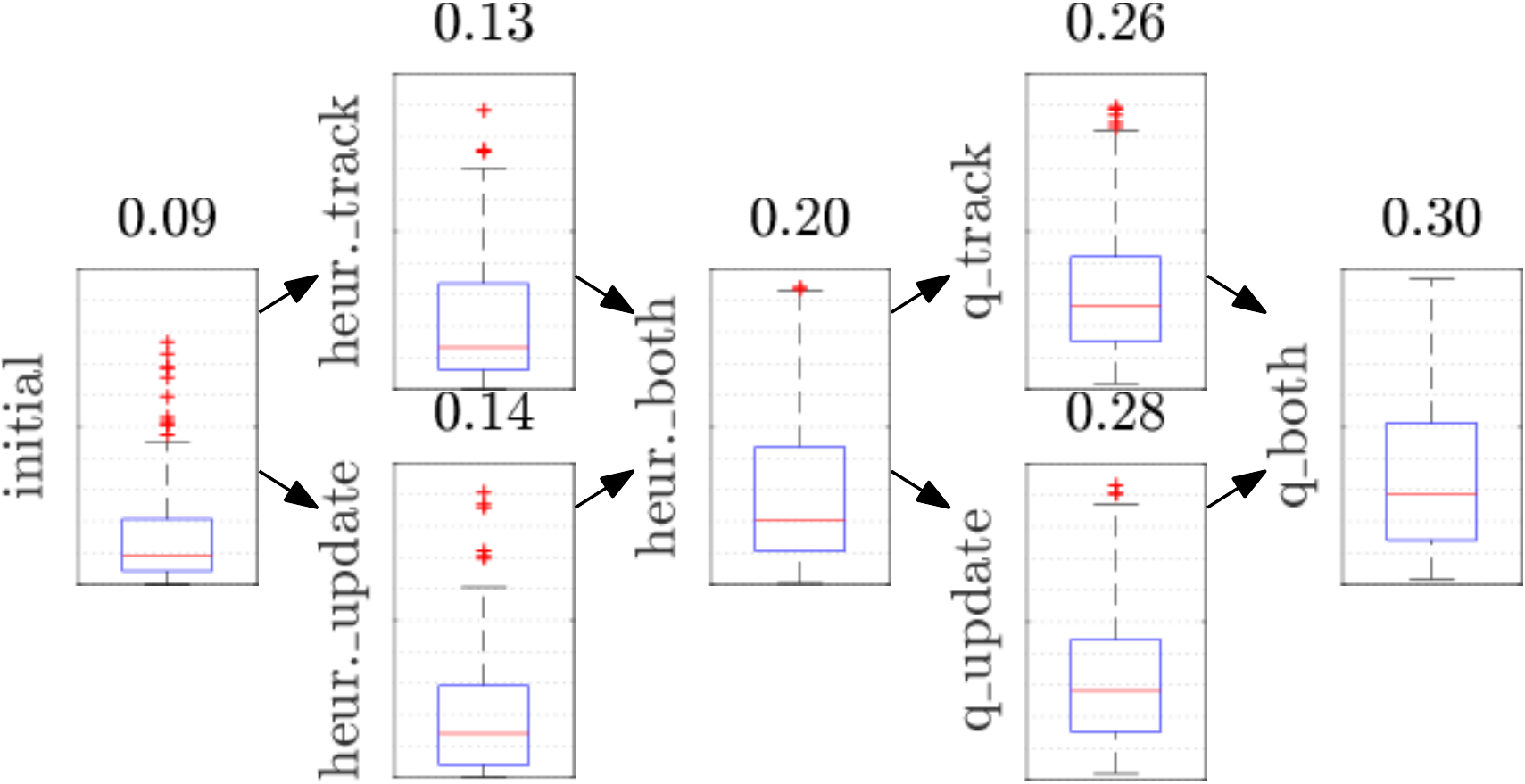}

\caption[Diagnostic experiments]{\textbf{System diagnostics: } 
Beginning with the \emph{initial} policy of FCNT~\cite{Wang2015}, 
we evolve towards our final data-driven policy objective. 
As shown, each component of our objective measurably improves performance
on the hold-out test-data. 
See Sec.~\ref{subsec:systemevolution} for discussion. }
\label{fig:ablative}
\end{figure}

{\bf Long-term videos:}
Qualitatively speaking, long term videos from the internet are much more difficult than standard benchmarks (\cf Fig~\ref{fig:qual_otb} and Fig.~\ref{fig:qual}).
First, many standard benchmarks tend to contain videos that are easy for most modern tracking approaches, implying that a method's rank is largely determined by performance on a small number of challenging videos. The easy videos focus on iconic~\cite{palmer1981canonical,berg2009finding} views with slow 
motion and stable lighting conditions~\cite{li2015nus}, featuring no
cuts or long-term occlusions~\cite{Maresca_2014_CVPR_Workshops}.  
Internet videos are significantly more complex. One major reason is
the presence of frequent cuts. 
We think that Internet videos with multiple cuts provide a valuable proxy for occlusions, particularly since they are a scalable data source.
In theory, long-term trackers must be able to re-detect the target
after an occlusion (or cut), but there is still much room for improvement. 
Also, many strange things happen in the wild world of Internet videos. 
For example, in \emph{Transform} the car transforms into a robot, confusing all trackers (including ours).
In \emph{SnowTank} the tracker must contend with many distractors (tanks of different colors and type) and widely varying viewpoint and scale. 
Meanwhile, \emph{JohnWick} contains
poor illumination, fast motion, and numerous distractors. 

\begin{figure}[tbp!]
    \centering
\resizebox{\linewidth}{!}{
\renewcommand{\tabcolsep}{0pt}
\def\arraystretch{0}

    \begin{tabular}{cccc}
    (a) Track \& Update & (b) Track Only & (c) Track Only & (d) Reinit \\
    \includegraphics[width=.33\linewidth,height=.32\linewidth]{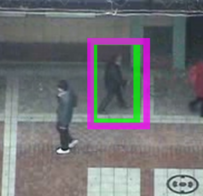} &
    \includegraphics[width=.33\linewidth,height=.32\linewidth]{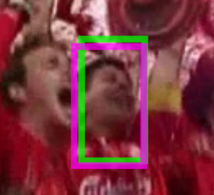} &
   \includegraphics[width=.33\linewidth,height=.32\linewidth]{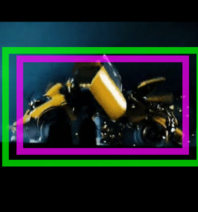} &    
    \includegraphics[width=.33\linewidth,height=.32\linewidth]{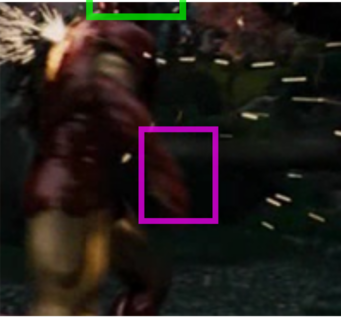} \\
    \includegraphics[width=.33\linewidth,height=.32\linewidth]{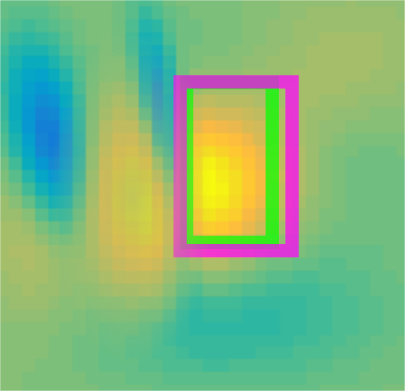} &
    \includegraphics[width=.33\linewidth,height=.32\linewidth]{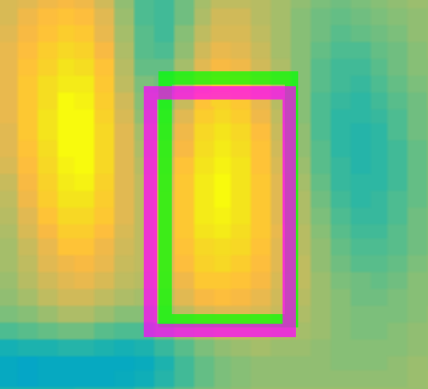} &
    \includegraphics[width=.33\linewidth,height=.32\linewidth]{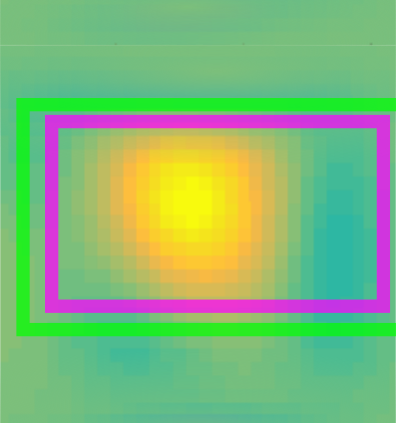} &    
    \includegraphics[width=.33\linewidth,height=.32\linewidth]{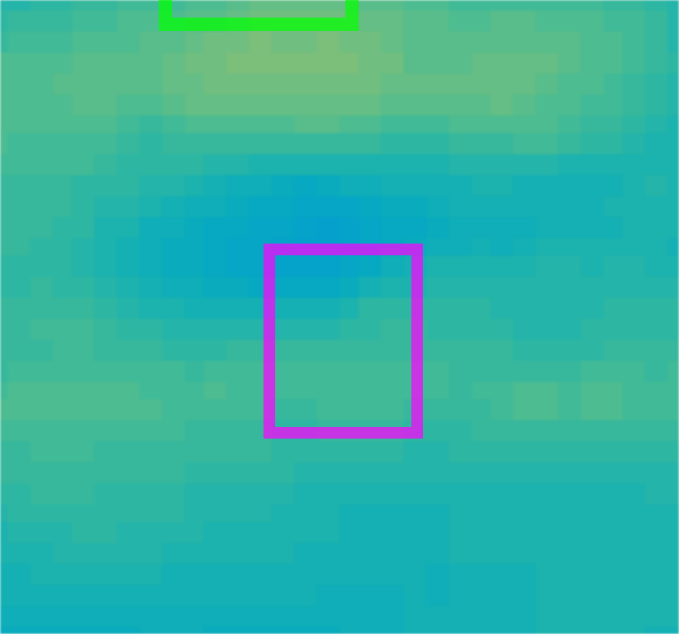} 
    \end{tabular}}

    \resizebox{\linewidth}{!}{
    \begin{tabular}{|c|c|c|c|}
        \hline
         Ground Truth & 
         \includegraphics[width=.33\linewidth,height=.05\linewidth]{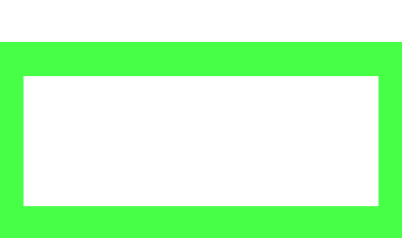} & 
         Tracker's Localization & 
         \includegraphics[width=.33\linewidth,height=.05\linewidth]{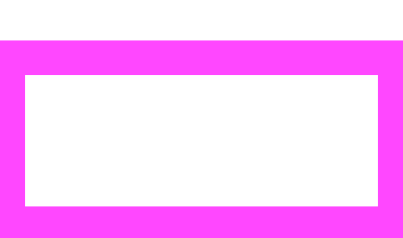} \\
         \hline
    \end{tabular}}
    
    \caption[Interpreting p-track's policy]{\textbf{What does p-track learn? } We show the actions taken by our tracker given four heatmaps. P-track learns to track and update appearance even in cluttered heatmaps with multiple modes ({\bf a}). However, if the confidence of other modes becomes high, p-track learns not to update appearance to avoid drift due to distractors ({\bf b}). If the target mode is heavily blurred, implying the target is difficult to localize (because of a transforming robot), p-track also avoids model update ({\bf c}). Finally, the lack of mode suggests p-track will reinitialize ({\bf d}).
    } 
    \label{fig:whatislearned}
\end{figure}

{\bf Long-term results: }  To evaluate results for ``in-the-wild'' long-term tracking,
we define a new 16-video held-out test set of long-term Internet videos that is {\em never used for training}. 
Each of our test videos contains at least 5,000 frames, a common definition of ``long-term''~\cite{supancic2013self,ma2015long,hong2015multi,Maresca_2014_CVPR_Workshops}.
We compare our method to various baselines in Fig.~\ref{fig:comparative}. 
Comparisons to FCNT~\cite{Wang2015} and CCOT~\cite{danelljan2016beyond} are particularly interesting since
we can make use of their TRACK and UPDATE modules. 
While FCNT performs quite well in the short-term (Fig.~\ref{fig:benchmarks}), 
it performs poorly on long-term sequences (Fig.~\ref{fig:comparative}).
However, by learning a policy for updating and reinitialization, we produce a state-of-the-art
long-term tracker. 
We visualize the learned policy in Fig.~\ref{fig:whatislearned}. 

\subsection{System Diagnostics}

\label{subsec:systemevolution}
We now provide a diagnostic analysis of various components of our system. We begin by examining several alternative strategies for making sequential decisions 
(Fig.~\ref{fig:ablative}).


\paragraph{FCNT vs. CCOT:} We use FCNT for our ablative analysis; initializing p-track using 
CCOT's more complex online heuristics proved difficult. However our final system
uses only our proposed offline heuristics, so we can  nonetheless train it using
CCOT's TRACK, REINIT, and UPDATE functions.
In Fig.~\ref{fig:comparative} we compare the final
p-trackers built using FCNT's functions against those built using
CCOT's functions. As consistent with prior work, we find that CCOT improves overall performance (from .30 to .36).

\paragraph{Online vs. offline heuristics:}  We begin by analyzing the online heuristic actions of our baseline tracker, FCNT. FCNT updates an appearance model when the predicted heatmap location is above a threshold, and always tracks without reinitialization. This produces a F1 score of .09. 
Next, we use offline heuristics to learn the best action to take. These correspond to tracking when the predicted object location is correct, and updating if the appearance model trained on the new patch produces higher scores for ground-truth locations. We train a classifier to predict these actions using the current heatmap. When this offline trained classifier is run at test time, F1 improves to .13 with the track heuristic alone, .14 with the update heuristic alone, and .20 if both are used.

\paragraph{Q-learning:} 
Finally, we use Q-learning to refine our heuristics (Eq.~\ref{eq:q}), noticeably improving the F1 score to .30. 
Learning the appearance update action seems to have the most significant effect on performance, producing an F1 score of .28 by itself.
During partial occlusions, the tracker learns to delicately balance
between appearance update and drift
while accepting a few failures 
to avoid the cost and risk of reinitialization. Overall, the learned policy dramatically outperforms the default online heuristics, {\bf tripling} the F1 score from 9\% to 30\%! 

\begin{figure}
    \centering
    \includegraphics[width=\linewidth]{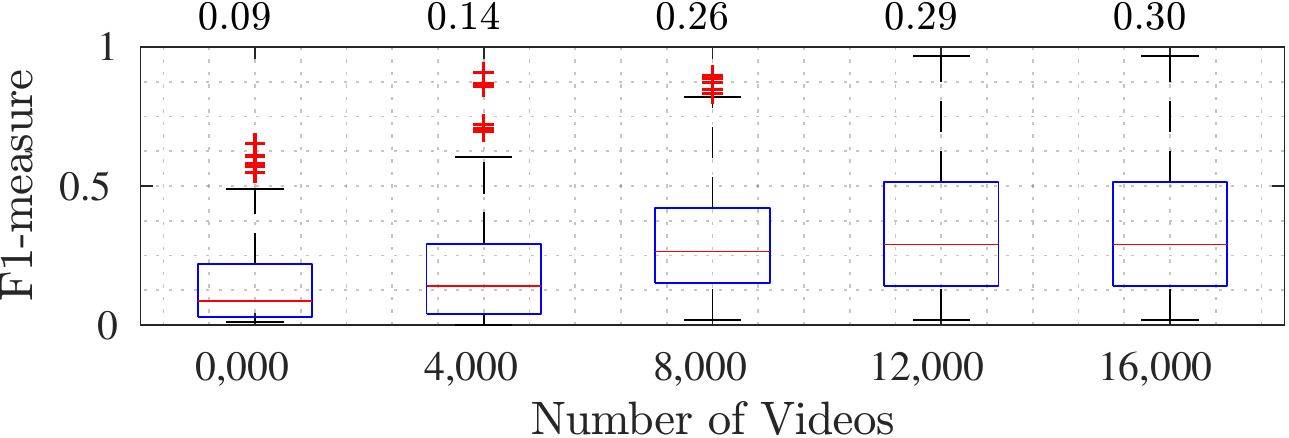}

    \caption[Training p-track]{\textbf{Our p-tracker's performance increases } as it learns its policy using additional Internet videos. Above, we plot the distribution of F1 scores on our hold-out test data, at various stages of training. At initialization, the average F1 score was $0.08$. After seeing 16,000 videos, it achieves an average F1 score of $0.30$. Our results suggest that {\em large-scale training}, made possible through interactive annotation, is crucial for learning good decision policies.}
\label{fig:overepochs}
\end{figure}   

\paragraph{Training iterations: }
In theory, our tracker can be interactively trained on a never-ending stream. 
However, in our experiments, Q-learning appeared to converge after seeing between 8,000 and 12,000 videos. 
Thus, we choose to stop training after seeing 16,000 videos.
In Fig.~\ref{fig:overepochs}, we plot performance versus training
iteration. 

\paragraph{Computation: }
As mentioned previously, comparatively slower trackers typically perform better~\cite{Kristan2016a}. 
On a Tesla K40 GPU, our tracker runs at approximately 10 fps. 
While computationally similar to~\cite{Wang2015}, we add the ability to recover from tracking failures by reinitializing through detection. 
To do so, we learn an attention policy that efficiently balances tracking versus
reinitialization. 
Tracking is fast because only a small region of interest (ROI) need be searched.
Rather than searching over the whole image during reinitialization, 
we select a random ROI (which ensures that our trackers operate at a fixed frame rate). In practice, we find that target is typically found in $\approx 15$ frames.

\paragraph{Conclusions:}
We formulate tracking as a sequential decision-making problem, where a tracker must update its beliefs about the target, given noisy observations and a limited computational budget. While such decisions are typically made heuristically, we bring to bear tools from POMDPs and reinforcement learning to learn decision-making strategies in a data-driven way. Our framework allows trackers to learn action policies appropriate for different scenarios, including short-term and long-term tracking. One practical observation is that offline heuristics are an effective and efficient way to learn tracking policies, both by themselves and as a regularizer for Q-learning. Finally, we demonstrate that reinforcement learning can be used to leverage massive training datasets, which will likely be needed for further progress in data-driven tracking.

\label{sec:conclusion}


{\small
\bibliographystyle{ieee}
\bibliography{egbib}
}

\end{document}